\documentclass[letterpaper, 10 pt, conference]{ieeeconf}  

\IEEEoverridecommandlockouts                              

\let\labelindent\relax

\usepackage{times}
\usepackage{latexsym}
\usepackage[T1]{fontenc}
\usepackage[utf8]{inputenc}

\usepackage{booktabs}

\usepackage{epsfig}
\usepackage{graphicx}
\usepackage{amsmath}
\usepackage{amssymb}
\usepackage{enumitem}
\usepackage{tabularx}
\usepackage{multirow}
\usepackage{epsfig}
\usepackage[dvipsnames]{xcolor}

\graphicspath{ {./images/} }
\usepackage[rightcaption]{sidecap}
\usepackage[export]{adjustbox}
\usepackage{wrapfig}
\usepackage{microtype}
\newcommand{\etal}{\textit{et al.}}
\usepackage{inconsolata}

\title{\LARGE \bf Ground then Navigate: Language-guided Navigation in Dynamic Scenes}

\author{Kanishk Jain$^{*}$, Varun Chhangani$^{*}$, Amogh Tiwari, K. Madhava Krishna and Vineet Gandhi
\thanks{*Equal Contribution}
\thanks{The authors are with Kohli Center on Intelligent Systems,  International Institute of Information Technology, Hyderabad, India, 500032.
        {\tt\small kanishk.j@research.iiit.ac.in}}%
\thanks{The work was supported in part by a Qualcomm Innovation Fellowship}%
}

\begin{document}

\maketitle
\thispagestyle{empty}
\pagestyle{empty}

\begin{abstract}

We investigate the Vision-and-Language Navigation (VLN) problem in the context of autonomous driving in outdoor settings. We solve the problem by explicitly grounding the navigable regions corresponding to the textual command. At each timestamp, the model predicts a segmentation mask corresponding to the intermediate or the final navigable region. Our work contrasts with existing efforts in VLN, which pose this task as a node selection problem, given a discrete connected graph corresponding to the environment. We do not assume the availability of such a discretised map. Our work moves towards continuity in action space, provides interpretability through visual feedback and allows VLN on commands requiring finer manoeuvres like "park between the two cars". Furthermore, we propose a novel meta-dataset CARLA-NAV to allow efficient training and validation. The dataset comprises pre-recorded training sequences and a live environment for validation and testing. We provide extensive qualitative and quantitive empirical results to validate the efficacy of the proposed approach.

\end{abstract}

\section{INTRODUCTION}

Humans have exceptional navigational abilities, which, combined with their visual and linguistic prowess, allow them to perform navigation based on the linguistic description of the objects of interest in the environment. This is in direct consequence of the human capability to associate visual elements with their linguistic descriptions. Referring Expression Comprehension (REC)~\cite{rohrbach2016grounding} and Referring Image Segmentation (RIS)~\cite{hu2016segmentation} are two tasks for associating the visual objects based on their linguistic descriptions using bounding-box-based and pixel-based localizations, respectively. However, it is non-trivial to utilize these localizations directly for a navigation task~\cite{deruyttere2019talk2car}. For example, consider the linguistic command "take a right turn from the intersection," an object-based localization is not usable for navigation as it does not answer the question "which region" on the road to navigate to. To solve the aforementioned issue, the task of Referring Navigable Regions (RNR) was proposed in \cite{Rufus_2021} to localize the navigable regions in a static front camera image on the road corresponding to the linguistic command. 

\begin{figure}[!t]
    \centering
    \begin{tabular}[t]{p{0.46\textwidth} }
        \includegraphics[width=0.46\textwidth]{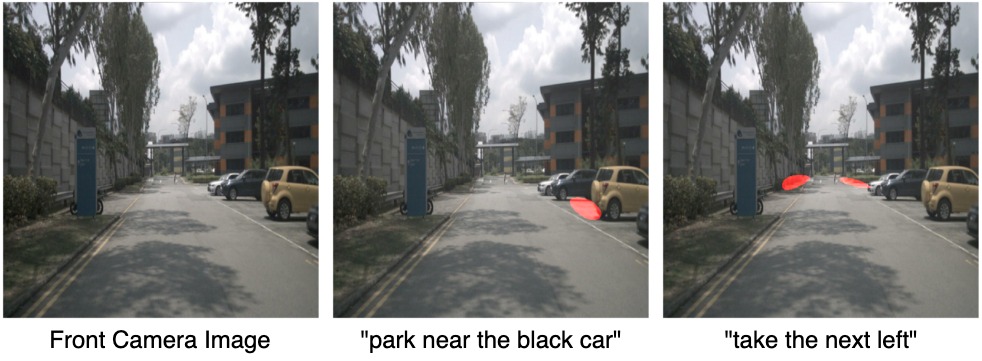} \\
    \end{tabular}
    \caption{A major limitation of single image based grounding methods is that they fail if the language command is not immediately visible, which restricts these methods to be used for VLN. Here, we show an example result using the RNR model and on an image from their dataset~\cite{Rufus_2021}. The model accurately localizes the black car (middle), however, it completely confuses when asked to predict the left turn, which does not exist in the current view.}
    \label{fig:teaser}
\end{figure}

Although these single image-based visual grounding methods~\cite{Rufus_2021,hu2016segmentation,deruyttere2019talk2car} showcase excellent ability of neural models to correlate visual and linguistic data, they are limited in many ways. These methods are trained on carefully paired data, assuming that the region to be grounded is always visible in the frame. They give an erroneous output when the region to be grounded is not currently visible, is occluded, or goes out of frame (as the carrier moves). Such scenarios are part of everyday language-guided navigation; for instance, consider a command ``Take a right once you see the traffic signal" the traffic signal here may not be immediately visible. Figure \ref{fig:teaser} illustrates an example, where the single image-based RNR method gives incorrect output, uncorrelated with the linguistic command. As a second major limitation, single image-based predictions~\cite{Rufus_2021} are devoid of any temporal context (short term or long term), which is crucial in successful navigation, especially in a dynamically changing environment. Finally, since single image methods are evaluated on frames from pre-recorded videos, they cannot be validated appropriately on their ability to complete the entire episode (from start to desired finish). Our work addresses these limitations and re-formulates the RNR approach to perform language-guided navigation in a dynamically changing environment by grounding intermediate navigable regions when the referred navigable region is not visible. 

Our work also contrasts with the prior art for language-guided navigation in both indoor and outdoor environments. Most existing works on indoor navigation~\cite{anderson2018vision,https://doi.org/10.48550/arxiv.2001.02330,zang-etal-2018-translating} assume that the navigational environment is fully known. This allows them to discretize the known map into a graphical representation, where the nodes are the set of navigable regions (landmarks) that the agent can navigate given the linguistic command. However, such an approach is not practical for outdoor settings (as studied in our work) where the environment is unknown. Moreover, even if the environment is known, discretization of the maps is not feasible when more refined localization and manoeuvres are required (e.g. ``stop beside the person with a red cap"). 

Finer control remains a challenge in indoor and outdoor VLN methods, which model navigation as a selection from a set of discrete actions~\cite{schumann-riezler-2022-analyzing,zhu-etal-2021-multimodal,xiang-etal-2020-learning} or as a reinforcement learning problem \cite{anderson2018vision,fu2019language}. For instance, one of the commonly used Touchdown dataset~\cite{chen2019touchdown} consists of pre-recorded google street view images and allows navigation across street views by choosing from a set of four discrete actions, i.e. FORWARD, RIGHT, LEFT and STOP. Discretizing the action space (and the environment) limits the type of navigational manoeuvres that can be performed. For instance, these methods~\cite{schumann-riezler-2022-analyzing,zhu-etal-2021-multimodal,xiang-etal-2020-learning} cannot be used for commands like "park between the two cars on the right", requiring fine-grained control.

\begin{figure*}[h]
\centering
\includegraphics[width=1.0\textwidth]{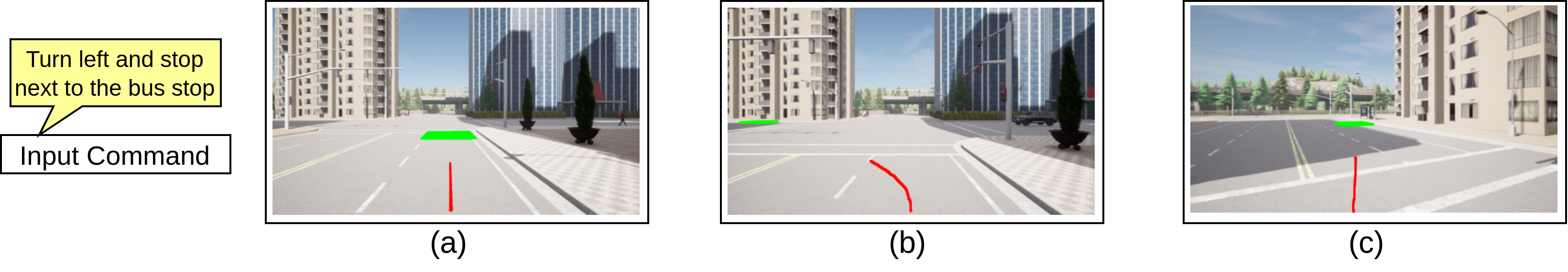}
\caption{Ground truth annotations of three sampled frames from an episode of the CARLA-NAV dataset. The textual command for the episode is shown on the top. The green rectangle illustrates the navigable region, and the red curve corresponds to the short future trajectory. (a) At the start, the left turn or the final navigable region is not visible, so a straight path is chosen as the intermediate mask; (b) the intermediate mask corresponding to the left turn; (c) the final navigable region (stop next to bus stop).}
\label{dataset_sample}
\end{figure*}

The aforementioned issues become apparent in a dynamically changing environment, where fine-grained control of the car's navigation and a fully navigable environment is required to adapt to the dynamic surroundings and perform navigational manoeuvres based on the linguistic command. In this paper, we present a novel meta-dataset in the CARLA environment~\cite{dosovitskiy2017carla} for outdoor navigation, which addresses the limitations associated with the existing navigation datasets. Additionally, the visual grounding-based approach combined with a planner allows us to have a fine-grained control over the vehicle as it enables navigation to any drivable region on the road.

Another concern with the previous methods~\cite{schumann-riezler-2022-analyzing,zhu-etal-2021-multimodal,anderson2018vision} is that their predictions are not human interpretable. It is non-trivial to understand their predictions as there is no feedback. Instead, in the visual-grounding-based approach, there is visual feedback associated with each prediction in terms of the segmentation mask corresponding to the navigable region on the road. We take a step forward and also predict the short term intermediate trajectories, using a novel multi-task network. Moreover, we perform live inference on the proposed meta-dataset in a dynamic environment. To the best of our knowledge, this is the first attempt towards live language-based navigation in outdoor environments. Overall, our paper makes following contributions:



\begin{itemize}
\item We present a vision language navigation tool, CARLA-NAV, on the CARLA simulator which provides fine-tuned control of the vehicle to execute various language-based navigational manoeuvres.
\item We propose a novel multi-task network for trajectory prediction and per-frame RNR tasks in dynamic outdoor environments. The prediction for each task is explainable and interpretable in the form of segmentation masks.
\item We perform real-time navigation in the CARLA environment with a diverse set of linguistic commands.
\item Finally, extensive qualitative and quantitative ablations are performed to validate the practicality of our approach.  

\end{itemize}

\section{RELATED WORK}
\subsection{Visual Grounding}

Visual grounding aims to help associate the linguistic description of entities with their visual counterparts by localizing them visually. There are two prevalent approaches for visual grounding: (a) proposal based and (b) segmentation based. Proposal-based grounding is formally referred to as Referring Expression Comprehension (REC). Most methods in REC follow a propose-then-rank strategy, where the ranking is done using similarity scores \cite{rohrbach2016grounding,hu2016natural,plummer2018conditional,rufus2020cosine} or through attention-based methods \cite{deng2018visual,zhang2018grounding,yang2019dynamic,qiu2020language}. The other approach is to localize the objects by their pixel-level segmentation mask, formally known as Referring Image Segmentation (RIS). In RIS, methods use different strategies to fuse the spatial information of the image with the word-level information of the language query \cite{shi2018key,ye2019cross,huang2020referring,jain2021comprehensive}. Recently, \cite{Rufus_2021} proposed the Referring Navigable Region (RNR) task to directly localize the navigable regions on the road corresponding to the language commands. However, their work limits to predictions on static images in pre-recorded video sequences~\cite{caesar2020nuscenes}. We propose reformulating the RNR task for dynamic outdoor settings and performing real-time navigation based on language commands.

\subsection{Language-based Navigation}

Majority of efforts on Vision Language Navigation (VLN) have focused on the indoor scenario. Availability of interactive synthetic environments has played a key role in indoor navigation research. The environments are either designed manually by 3D artists~\cite{kolve2017ai2,wu2018building} or are constructed using RGB-D scans of actual buildings~\cite{chang2017matterport3d}. Existing methods have approached language guided navigation in variety of ways, including imitation learning~\cite{nguyen2019vision,nguyen2019help}, behavior cloning \cite{das2018neural}, sequence-to-sequence translation \cite{anderson2018vision} and cross-modal attention~\cite{cornia2019perceive}. In these methods, the navigation is modelled as traversing an undirected graph, presuming known environment topologies. In recent work, \cite{krantz2020beyond} suggest that the performance in prior `navigation-graph' settings may be inflated by strong implicit assumptions. Hu~\etal~\cite{hu2019you} questions the role of visual grounding itself by highlighting that models which only use route structure outperform their visual counterparts in unseen new environments. Most indoor VLN methods are also hindered by limiting the output to a discretized action space~\cite{irshad2021hierarchical}.

For outdoor VLN, Sriram~\etal~\cite{sriram2019talk} use CARLA environment to perform navigation as waypoint selection problem, however, their work limits to only turning actions. The Talk2Car dataset limits to localizing the referred object~\cite{deruyttere2019talk2car}. Another line of work focuses on interactive navigation environment of Google Street View~\cite{mirowski2018learning}. The Touchdown dataset~\cite{chen2019touchdown} proposes a task of following instructions to reach a goal (identifying a hidden teddy bear). Map2Seq dataset~\cite{schumann2020generating} learns to generate navigation instructions that contain visible and salient landmarks from human natural language instructions. The navigation on both datasets is modelled as node selection in a discrete connectivity graph. Most methods~\cite{schumann-riezler-2022-analyzing,zhu-etal-2021-multimodal,xiang-etal-2020-learning} using these datasets, solve outdoor VLN as sequence to sequence translation in a discrete action space. The role of vision modality remains illusive when tested in unseen areas~\cite{schumann-riezler-2022-analyzing}. In this work, we propose a paradigm shift towards utilizing RNR-based approaches for VLN. The explicit visual grounding forces the network to utilize visual information. Integrating with a local planner, the navigation is performed in a continuous space, without any reliance on the map information.

\section{Dataset}


The proposed CARLA-NAV dataset was curated using the open-source CARLA Simulator for autonomous driving research. It contains episodic level data, where each episode consists of a language command and the corresponding video from CARLA Simulator of navigation towards the final goal region described by the command. Example ground truth annotations from an episode from the CARLA-NAV dataset are shown in Figure~\ref{dataset_sample}. The ground truth segmentation mask for each frame either corresponds to the final or an intermediate navigable region. Each frame is additionally annotated with a plausible future trajectory of the vehicle in the next few frames. 
%


The dataset includes video sequences captured in 8 different maps, 14 distinct weather conditions, and a diverse range of dynamically moving vehicles and passengers in the environment. The language commands in our dataset contain detailed visual descriptions of the environment and describe a wide range of manoeuvres. Commands can either have a single manoeuvre (e.g. park behind the black car on the left) or multiple manoeuvres (e.g. take a right turn and park near the bus stand). A maximum of three manoeuvres are included in the dataset. The dataset statistics for the CARLA-NAV dataset are illustrated in Table \ref{table:data_stats}. Since, each episode contains multiple frames, the overall dataset size is 83,297 frames for all the splits. During data collection, in each episode, the vehicle is spawned in a randomly selected map at a random position. During the training phase, we use the pre-recorded sequence for the network training. However, during the inference phase on validation and test splits, for each episode, we spawn the vehicle at the corresponding starting location, and the navigation is performed live based on network prediction and not on the pre-recorded sequences.


\subsection{Dataset Creation}

We created a data-collection toolkit on top of Carla's API and plan to open-source it upon acceptance. The data collection process involves an observer, a navigator and a verifier. The annotation happens in a three step manual process: (a) observer: providing a language command, (b) navigator: navigating the Carla environment through mouse clicks corresponding to navigable regions and (c) verifier: verifying the recorded episode for inclusion in dataset, if it correctly maneuvers corresponding to the given command. The navigator is provided with a "restart" option to restart the navigation in case of erroneous clicks. 

%
\begin{table}[t]
\begin{center}
\scriptsize
\begin{tabular}{|c|c|l|c|c|}
\hline
split & \# episodes & \# frames & \begin{tabular}[c]{@{}c@{}}command length\\ (words)\end{tabular} & clicks \\ \hline
train & 500 & 75,010 & 6.92 & 1.94 \\ \hline
val   & 25  & 3,300  & 6.76 & 2.00 \\ \hline
test  & 34  & 4,987  & 7.44 & 1.97 \\ \hline
\end{tabular}
\end{center}
\caption{Dataset statistics for the CARLA-NAV dataset. Except "\# episodes" and "\# frames", average values per episode are reported for the other columns.}
\label{table:data_stats}
\end{table}
%
 The 2D point corresponding to the mouse click in the front view of the car is transformed into the 3D world coordinates using Inverse Projective transform; and this 3D position is passed as input to the local planner to navigate the CARLA environment. We use CARLA's default planner for our case; however owing to the modular design, in future it can be replaced by any state-of-the-art planner.
 
 
 An episode comprise of multiple mouse clicks, until the final navigable region is not visible in the front view. These intermediate mouse clicks signify the intermediate navigable regions, and the last mouse click depicts the final goal region corresponding to the command. The mouse clicks are converted into segmentation masks by drawing a 3m $\times$ 4m rectangle (approx size of a car) in the top view, centered at the mouse click. The rectangle is then projected in the front-camera.  Overall, only the language commands and mouse clicks require manual effort, the rest of the annotation process is fully automated. 
 
Furthermore, for allowing actionable intervention we predict and visualize the future short term trajectory of the vehicle. For the trajectory prediction task, we take the 3D position of the vehicle in the successive frames and project the 3D positions to the front-camera image using the projective transformation. We treat trajectory prediction as a dense prediction task. The task is meant for added human interpretability and is not quantitatively evaluated.


\begin{figure*}[t]
\centering
\includegraphics[width=1.0\textwidth]{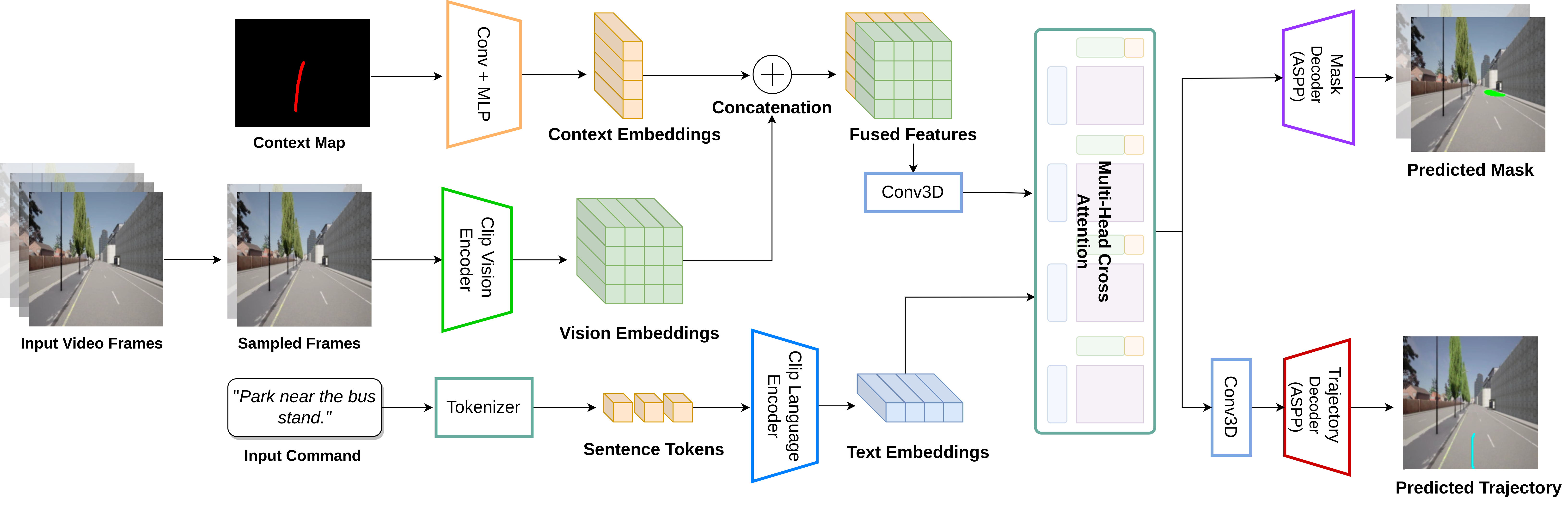}
\caption{Overall pipeline of the proposed approach. Given the visual frames, already executed past trajectory (context map) and the textual command, the network predicts a segmentation map corresponding to the navigable region and a plausible future trajectory.}
\label{proposed_pipeline}
\end{figure*}

\section{Problem Statement}

Given an input of video frames $V=\{v_{t-k}, v_{t-k+1}, ... v_{t}\}$, contextual historical trajectory $P$ and a language command $L = \{l_1, l_2, ...l_N\}$, where $t$ is the current timestamp, $k$ is the window size for historical frames and $N$ is the maximum number of words in the linguistic expression, the goal is to predict the navigable mask $y_t$ and the future trajectory mask $z_t$ corresponding to the frame from current timestamp, i.e. $v_t$. The contextual trajectory $P$ is utilized to ensure that the network gets the contextual information necessary to identify which part of the linguistic command has been executed. For example, if the linguistic command is "turn left and park near the blue dustbin", the contextual trajectory will provide information regarding the trajectory already taken by the vehicle, i.e. whether the "left turn" has been taken or not. The spatial location of the navigation mask should determine the trajectory path's direction; similarly, the orientation of the trajectory path should determine the location of the navigable regions. In the next section we describe the network architecture and the training process.

\section{Methodology}

We propose a novel multi-task network for navigation region prediction and future trajectory prediction tasks. Both tasks are treated as dense prediction tasks to make them interpretable for practical scenarios. We convert the dense pixel points to 3D world coordinates using inverse projective transformation during real-time inference. The architecture for our model is illustrated in Figure~\ref{proposed_pipeline}. In this section, we  describe the feature extraction process and the architecture in detail.

We utilize CLIP \cite{Radford2021LearningTV} to extract both linguistic and visual features. For the linguistic expression $L = \{l_1, l_2, ...l_N\}$, where $l_i$ is the $i^{th}$ word of the expression, we tokenize the linguistic command using CLIP tokenizer and pass it through CLIP architecture to compute word-level feature representation $F^l = \{f^{l}_{1}, f^{l}_{2}, ..., f^{l}_{N}\}$ of shape $\mathbb R^{N \times C_l}$. For visual frames $V=\{v_{t-k}, v_{t-k+1}, ... v_{t}\}$, the CLIP architecture encodes the video features as $F^v=\{f^{v}_{t-k}, f^{v}_{t-k+1}, ... f^{v}_{t}\}$. Finally, for the trajectory context $P$, we project the past trajectory on an image having same size as the input video frames $v_t$'s and pass it through convolution and a MLP layer to get feature map $F^p$ with the same feature size as video frame features $f^{v}_{t}$'s.

The input to our network are the video frames $V$, historical trajectory context $P$ and the language command $L$. Specifically, for video $V$, we get visual features $F^v$ of shape $ \mathbb R^{C_v \times T \times HW}$, where H, W, T, and C represent the height, width, time, and channel dimensions, respectively. The trajectory context feature $F^p \in \mathbb R^{C_v \times 1 \times HW}$ contains information about the past trajectory taken by the vehicle. Following feature extraction, we concatenate the trajectory context feature $F^p$ with video features $F^v$ along the temporal dimension resulting in joint feature $F^{vp} \in \mathbb R^{C_v \times (T+1) \times HW}$ capturing the video and trajectory related contextual information. Finally, we apply multi-head self-attention over the joint contextual feature $F^{vp}$ and linguistic feature $F^l$ in the following manner,
\begin{equation} \label{eq1}
\begin{split}
    F &= F^{vp} \odot F^l \\
    A &= \text{Mhead}(F, F, F)\\
    M &= \text{Conv3D}(A \ast F)
\end{split}
\end{equation}

Here, $\odot$ represents the length-wise concatenation of the word-level linguistic features $F^l$ and the joint feature $F^{vp}$, $\text{Mhead}$ is the multi-head self-attention over the multi-modal features $F$ and $\ast$ represents the matrix multiplication. $\text{Conv3D}$ represents 3D convolution operation and is used to collapse the temporal dimension, $M$ is the final multi-modal contextual feature with information from both visual and linguistic modalities.

Next, we describe the procedure for predicting the navigation and trajectory prediction masks. We want the future trajectory and the navigable region for the current time-step to be correlated with each other, i.e. the future trajectory should point in the direction of the predicted navigable region. Consequently, we utilize the multi-modal contextual feature $M$ to predict the segmentation masks corresponding to the navigation and trajectory prediction tasks. For each task, we have a separate segmentation head, where each segmentation head comprises of sequence of convolution layers with upsampling operation. For training the segmentation masks, we utilize combo loss \cite{taghanaki2019combo} which is a combination of binary cross-entropy loss and dice loss:
\begin{equation} \label{eq2}
\begin{split}
    L_{bce} &= -{(y_{t}\log(\hat{y_{t}}) + (1 - y_{t})\log(1 - \hat{y_{t}}))}\\
    L_{dice} &= 2 * \frac{\hat{y_{t}} \cap y_{t}}{\Sigma{\hat{y_{t}}} + \Sigma{y_{t}}}\\        L_{combo} &= \lambda L_{bce} - (1 - \lambda) L_{dice}
\end{split}
\raisetag{3\normalbaselineskip}
\end{equation}
The proposed approach is end-to-end trainable and the predicted trajectory is highly correlated with the predicted navigation mask, as a result the predicted trajectory is interpretable in the sense that it suggests the future route to be taken by the autonomous vehicle.


\section{Experiments}

\textbf{Implementation Details: }We utilize CLIP backbone \cite{Radford2021LearningTV} for feature extraction. The frames are selected with a stride of $10$ and are resized to $224 \times 224$ resolution. After feature extraction, we get per-frame visual features of spatial resolution $H=W=7$ and channel dimension $C_v=512$. For the historical contextual trajectory, we plot the trajectory from the starting location of episode to the current timestamp and resize it to $640 \times 480$ spatial resolution image, this is passed as input through convolution + MLP layers to obtain trajectory features with same resolution as per-frame visual features. For linguistic features, we use the CLIP tokenizer followed by the CLIP language encoder to compute the word-level features corresponding to the linguistic command. Maximum length of command is set to $N=20$ and the channel dimension is $C_l=512$. We use batch size of 32 and our network is trained using AdamW optimizer, the initial learning rate is set to $1e^{-4}$ and polynomial learning rate decay with power of 0.5 is used. For the combo loss, we set $\lambda = 0.3$. All the methods and baselines were trained from scratch using the CARLA-NAV dataset. For the single frame methods~\cite{Rufus_2021}, individual image and language pairs were used for training (with and without context map).  

\textbf{Live-Navigation: }In order to utilize the segmentation mask corresponding to the navigable region directly for navigation, we first need to sample a point from the predicted region. We take the largest connected component from the predicted mask and use its centroid as the target point for the local planner. As we move closer to the final navigable region, the distance between the current car location and the centroid target location consistently decreases. Simultaneously, the area of the predicted mask should increase as we move closer to the target region due to the perspective viewpoint of the front camera. Consequently, we use an area-based threshold to determine if the predicted navigation mask corresponds to the final navigable region or not. If the area of the predicted navigation mask is larger than the threshold for five consecutive times, we treat the predicted region as the final goal region corresponding to the linguistic command and stop the navigation.

\textbf{Evaluation Metrics:} Like previous approaches to VLN \cite{schumann-riezler-2022-analyzing, xiang-etal-2020-learning, chen2019touchdown}, we use the gold standard \textbf{\textit{Task Completion}} metric to measure the success ratio for the navigation task. In addition, we use \textbf{\textit{Frechet Distance}} \cite{eifer1994frechet} and \textbf{\textit{normalized Dynamic Time Warping (nDTW)}} \cite{ilharco2019general} metrics to compare the predicted navigation path during live inference with the ground truth navigation path. 

\subsection{Experimental Results}

\begin{table}[]
\begin{center}
\small
\begin{tabular}{|c|cc|}
\hline
\multirow{2}{*}{Method} & \multicolumn{2}{c|}{Task Completion}               \\ \cline{2-3} 
                        & \multicolumn{1}{c|}{Val}           & Test          \\ \hline
RNR-S                   & \multicolumn{1}{c|}{0.44}          & 0.29          \\ \hline
RNR-SC                  & \multicolumn{1}{c|}{0.52}          & 0.32          \\ \hline
CLIP-S                  & \multicolumn{1}{c|}{0.48}          & 0.47          \\ \hline
CLIP-SC                 & \multicolumn{1}{c|}{0.52}          & 0.50          \\ \hline
CLIP-M                  & \multicolumn{1}{c|}{0.56}          & 0.55          \\ \hline
CLIP-MC                 & \multicolumn{1}{c|}{\textbf{0.72}} & \textbf{0.68} \\ \hline
\end{tabular}
\end{center}
\caption{Results on the \textit{Task Completion} metric. The superior performance of proposed approach CLIP-MC, showcases the effectiveness of historical context for the navigation task.}
\label{table:table_1}
\end{table}

We compare our proposed approach CLIP-MC against the RNR-based approach proposed in \cite{Rufus_2021}. We use their proposed approach with CLIP-based backbone, CLIP-S as the baseline for our experimental results. The original RNR approach is limited to using a static scene with linguistic commands for navigation, which fails in a dynamically changing environment where the scene can change drastically when we start the navigation. Additionally, we motivate the benefits of contextual trajectory and multiple frames by presenting two variant baselines, (1) multiple frames without a contextual trajectory CLIP-M and (2) single frame with contextual trajectory CLIP-SC. Table \ref{table:table_1} presents the results on the gold standard \textit{Task Completion} metric and Table \ref{table:table_2} presents the results on \textit{Frechet Distance} and \textit{nDTW} metrics.

We observe that our proposed approach CLIP-MC outperforms all the other variants. Introducing historical contextual trajectory consistently helps improve performance as it increases by $4$\% and $16$\% in cases of single-frame approaches (CLIP-SC, CLIP-S) and multi-frame approaches (CLIP-MC, CLIP-M), respectively on the validation split. Furthermore, the multi-frame approach CLIP-M gives an improvement of $8\%$ on both the validation and test splits, respectively, over the single-frame approach CLIP-S. These results indicate that a combination of multiple frames and contextual trajectory are required to effectively tackle the VLN task.

In Table \ref{table:table_2}, we present experimental results on the \textit{Frechet Distance} and \textit{nDTW} metrics. Our reformulated approach CLIP-MC outperforms all other variants by significant margins. However, we would like to stress that these metrics are not robust indicators of the average performance on the actual navigation task, as a single outlier can drastically affect the final score. For example, on a single instance if "a left turn" is taken instead of "a right turn," the predicted trajectory will diverge from the ground truth trajectory, and will lead to significant deviation in the average scores.

\textbf{Effect of feature extraction backbone: }Additionally, we compare our CLIP-based single frame approaches with the original non-CLIP RNR approach proposed in \cite{Rufus_2021}, referred to as RNR-S and RNR-SC (single frame without and with context, respectively) in Table \ref{table:table_1}. Both RNR-S and RNR-SC are trained from scratch on the proposed CARLA-NAV dataset. The results showcase the advantage of superior multi-modal features captured by the CLIP-based approaches over non-CLIP approaches, as the performance consistently increases on the challenging test split in case of both with context (RNR-SC, CLIP-SC) and without context (RNR-S, CLIP-S).

\begin{table}[]
\begin{center}
\small
\begin{tabular}{|c|cc|cc|}
\hline
\multirow{2}{*}{Method} & \multicolumn{2}{c|}{Frechet Distance $\downarrow$}                & \multicolumn{2}{c|}{nDTW $\uparrow$}                          \\ \cline{2-5} 
        & \multicolumn{1}{c|}{Val}   & Test  & \multicolumn{1}{c|}{Val}  & Test \\ \hline
RNR-S   & \multicolumn{1}{c|}{28.14} & 42.45 & \multicolumn{1}{c|}{0.35} & 0.16 \\ \hline
RNR-SC  & \multicolumn{1}{c|}{21.64} & 44.65 & \multicolumn{1}{c|}{0.45} & 0.33 \\ \hline
CLIP-S  & \multicolumn{1}{c|}{40.30} & 42.53 & \multicolumn{1}{c|}{0.23} & 0.24 \\ \hline
CLIP-SC & \multicolumn{1}{c|}{35.58} & 38.49 & \multicolumn{1}{c|}{0.36} & 0.39 \\ \hline
CLIP-M  & \multicolumn{1}{c|}{32.92} & 53.10 & \multicolumn{1}{c|}{0.39} & 0.26 \\ \hline
CLIP-MC                 & \multicolumn{1}{c|}{\textbf{13.54}} & \textbf{15.06} & \multicolumn{1}{c|}{\textbf{0.54}} & \textbf{0.59} \\ \hline
\end{tabular}
\end{center}
\caption{Experimental results on the \textit{Frechet Distance} and \textit{nDTW} metrics. $\downarrow$ indicates lower value is better and $\uparrow$ indicates that the higher value is better.}
\label{table:table_2}
\end{table}


\begin{table}[]
\begin{center}
\small
\begin{tabular}{|c|c|ccccc|}
\hline
\multirow{2}{*}{Method} & \multirow{2}{*}{Split} & \multicolumn{5}{c|}{Task Completion}                                                                                 \\ \cline{3-7} 
                        &                        & \multicolumn{1}{c|}{n=1}  & \multicolumn{1}{c|}{n=2}  & \multicolumn{1}{c|}{n=4}  & \multicolumn{1}{c|}{n=6}  & n=8  \\ \hline
\multirow{2}{*}{CLIP-MC} & val & \multicolumn{1}{c|}{0.52} & \multicolumn{1}{c|}{0.48} & \multicolumn{1}{c|}{0.52} & \multicolumn{1}{c|}{0.68} & 0.72 \\ \cline{2-7} 
                        & test                   & \multicolumn{1}{c|}{0.50} & \multicolumn{1}{c|}{0.53} & \multicolumn{1}{c|}{0.56} & \multicolumn{1}{c|}{0.62} & 0.68 \\ \hline
\multirow{2}{*}{CLIP-M}  & val & \multicolumn{1}{c|}{0.48} & \multicolumn{1}{c|}{0.44}     & \multicolumn{1}{c|}{0.48}     & \multicolumn{1}{c|}{0.52} & 0.56 \\ \cline{2-7} 
                        & test                   & \multicolumn{1}{c|}{0.47} & \multicolumn{1}{c|}{0.47}     & \multicolumn{1}{c|}{0.50}     & \multicolumn{1}{c|}{0.53} & 0.55 \\ \hline
\end{tabular}%
\end{center}
\caption{Ablation on the number of frames for multi-frame models for the Task Completion metric.}
\label{table:table_3}
\end{table}

\textbf{Effect of Number of Frames: }In Table \ref{table:table_3}, we study the impact of the number of video frames on the multi-frame models for the Task Completion metric. As the number of video frames increases, the visual modality's contextual information also increases. We hypothesize that the network should utilize this additional contextual information and employ it effectively for the VLN task. The results in Table \ref{table:table_3} indeed corroborate our hypothesis, as we observe consistent performance gains as the number of video frames increases. The networks with $n = 1$ frame give the same performance as the corresponding single-frame variants. We obtain the best performance with $n = 8$ frames with CLIP-MC on both validation and test splits.

\subsection{Qualitative Results}
\begin{figure}[t]
\centering
\includegraphics[width=0.43\textwidth]{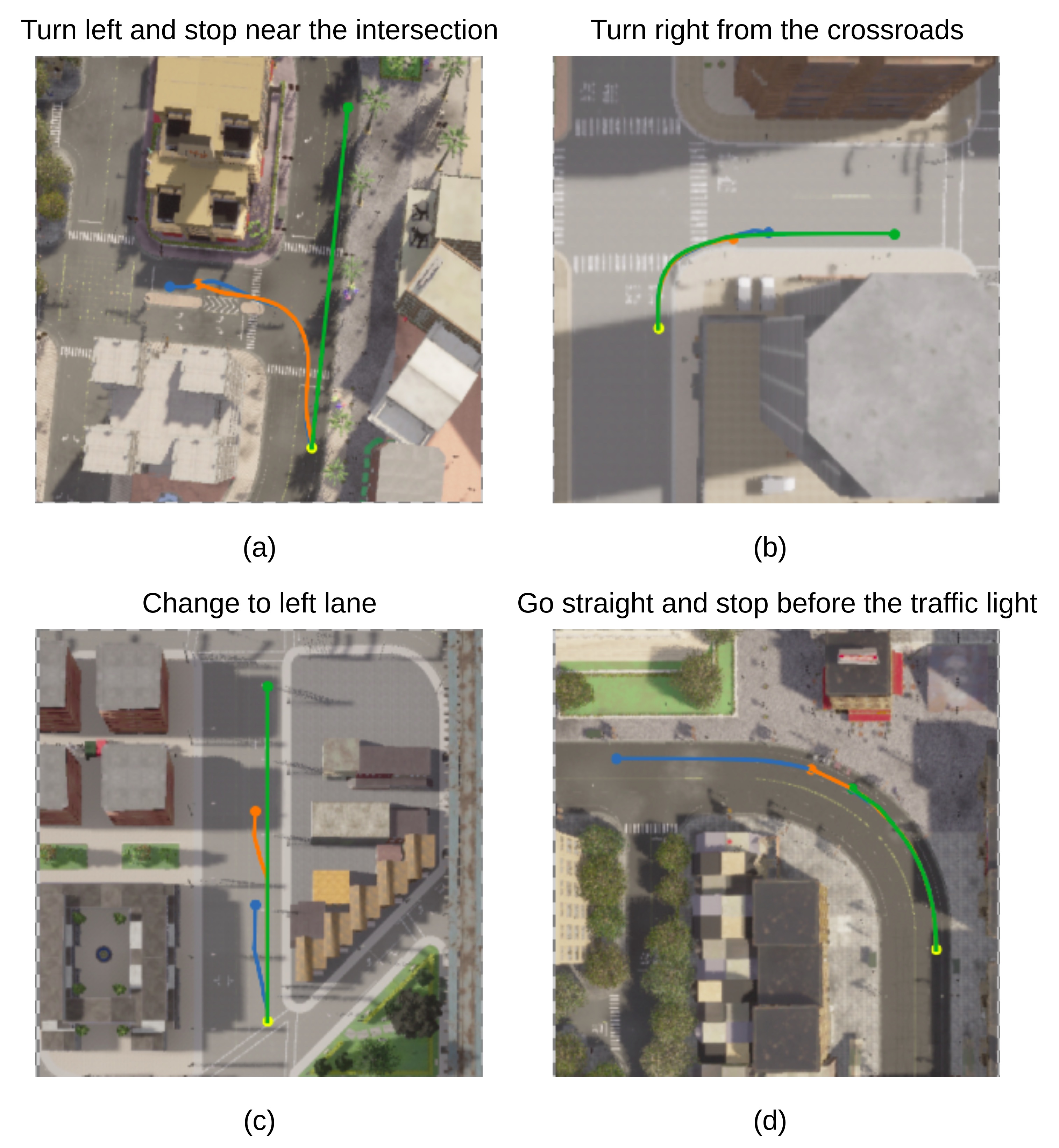}
\caption{Qualitative navigation results in the CARLA-NAV dataset. \textcolor{yellow}{Yellow} represents the starting point for the navigation. \textcolor{orange}{Orange} is used to depict the navigational path taken by CLIP-MC network, \textcolor{green}{green} denotes RNR-S network's navigational path and \textcolor{blue}{blue} represents the ground-truth path.}
\label{fig:qualitative_1}
\end{figure}

In Figure \ref{fig:qualitative_1}, we qualitatively compare the proposed approach CLIP-MC with the RNR approach (RNR-S) proposed in \cite{Rufus_2021}. We juxtapose the entire navigation path taken by each approach during live inference for a given linguistic command and overlay it on the aerial map of the CARLA environment. We showcase successful navigation scenarios of CLIP-MC in (a) , (b) and (c). With additional contextual information from multiple frames and historical trajectory, CLIP-MC can successfully perform "turning" and "stopping" based navigational manoeuvres. While the RNR approach, without any contextual information and trained on static images, fails. For the command ``change to the left lane", RNR-S fails to change the lane and continues in a straight line. While CLIP-MC manages to change the lane with a slight delay. For the example in the bottom-right corner, the road is curved in left direction and both the CLIP-MC and RNR-S stop much before the traffic light, as they mistake the curve with an intersection.

\section{Conclusion}

This paper proposes a language-guided navigation approach in dynamically changing outdoor environments. We reformulate the RNR approach, designed for static scenes to make it amenable for dynamic scenes. Our approach explicitly utilizes visual grounding directly for the navigation task. Along the same lines, we propose a novel meta-dataset CARLA-NAV, containing realistic scenarios of language-based navigation in dynamic outdoor environments. Additionally, we propose a novel multi-task grounding network for the tasks of navigable region and future trajectory prediction. The predicted navigable regions are explicitly used for navigating the vehicle in the dynamic environment. The predicted future trajectories bring interpretability to our approach and correlate with the predicted navigable region, i.e., they indicate the vehicle's navigational route. Furthermore, the proposed approach allows us to perform live navigation in a dynamic CARLA environment. Finally, quantitative and qualitative results validate our approach's effectiveness and practicality. Future work should explore domain adaptation techniques like \cite{kundu2021generalize, kang2020pixel} to ensure adaptability to real-world scenes and a learnable stopping criteria.



\bibliography{main}
\bibliographystyle{IEEEtran}

\end{document}